# Breast Cancer classification by adaptive weighted average ensemble of previously trained models

Mosab. S. M. Farea[1], Zhe Chen[2,*]

[1] College of Computer and Information, Hohai University, Nanjing 210098, China
[2] College of Computer and Information, Hohai University, Nanjing 210098, China

**Abstract**

**Breast cancer is a serious disease that inflicts millions of people each year, and the number of cases is increasing. Early detection is the best way to reduce the impact of the disease. Researchers have developed many techniques to detect breast cancer, including the use of histopathology images in CAD systems. This research proposes a technique that combine already fully trained model using adaptive average ensemble, this is different from the literature which uses average ensemble before training and the average ensemble is trained simultaneously. Our approach is different because it used adaptive average ensemble after training which has increased the performance of evaluation metrics. It averages the outputs of every trained model, and every model will have weight according to its accuracy. The accuracy in the adaptive weighted ensemble model has achieved 98% where the accuracy has increased by 1 percent which is better than the best participating model in the ensemble which was 97%. Also, it decreased the numbers of false positive and false negative and enhanced the performance metrics.**

**Keywords**

Transfer learning, Average ensemble, Breast cancer classification, CNN models, weighted average ensemble

## 1. Introduction

Cancer is a major global health problem which is considered one of the measure causes of death globally. According to WHO, breast cancer is the second largest cause of death worldwide, and it is reported in 2018 that almost 9.6 million death cases occurred due to the breast cancer around the globe. BC can affect women and men; however, it is mainly widespread in women who may develop this tragical disease in a period of her lifetime. The breast cancer institute stated that breast cancer is one of the death-leading diseases that plague women in the world[1]. Cancer is caused by the uncontrollable growth of the cells which cumulate to create tumor. Regularly, pathologists classified cancers into two types, one is benign and the other one is malignant. In a benign tumor case, the cells build up abnormally and create a lump. However, they do not invade other neighboring organs of the body so they are not classified as a cancer. Cancer begins as a benign type and, without a treatment at the early stages, it turns to become a malignant type. Malignant tumor cells are prone to invade the neighboring organs if no medical intervention is taken. For example, if a malignant tumor is not remedied, it may reach into the muscles under the breast, which is hard to remove and the endanger of recurrence is much higher. Early breast cancer disease detection and prevention enhance survival by 85 [2]. Pathologists use morphological abnormalities of the nucleus as the main feature to differentiate between malignant and benign cancerous [3] .early diagnosis helps the treatment to be more effective

---

[*] Corresponding author

[4]. there are many different tests that can be used to find and locate breast cancer. Some of these tests, such as magnetic resonance imaging MRI and computed tomography CT scans, mammogram, ultrasound and histopathological images. biopsies or Histopathological images are some of the first screening methods that serve to diagnose cancer or to see how far it has spread. These are used for those at danger of having breast cancer. These tests are interpreted by specialists, such as pathologists and radiologists. However, due to the very complex and there is a lot of information to process in images, it is also possible that even specialists can sometimes miss cancer cells on the images. Biopsy test is used as a technique for the diagnosis of the breast cancer which needs the experiences of the pathologists to diagnose the test, this task is always a time-consuming and in-depth assessment [5]. Many researchers developed several CAD systems for different diseases including bladder cancer, lung cancer, skin cancer, prostate cancer, colon cancer, cervical cancer, liver cancer and breast cancer [6][7][8][9][10][11]. Pathologists can use Computer aided detection (CAD) techniques to accelerate the process and to achieve early detection of breast cancer [12]. Artificial intelligence has improved the CAD systems through its several areas such as machine learning and deep learning. Machine learning contains four phases which are preprocessing, segmentation of the region of interest (ROI), features extraction which is a challenging task and selection, and lastly classification of suspicious lesions. The prediction accuracy of ML algorithms and their behavior is affected by the selections of features chosen [13]. The images are used as the building blocks of these systems to evaluate these images which can show breast-related information; however, the cancer signs are very subtle and their different formations reveal at their early stages [14]. Another area of artificial intelligence is deep learning which uses neural network that is inspired by biological neurons, it is composed of interconnected neurons which process the data through adjusting the weights and biases. Deep learning differs from machine learning because it does not need hand-crafted features and of its abilities to learn from complex image features, deep learning can be trained on huge datasets. These techniques which teaches the computer to learn and do tasks that need abilities of smart creatures are used in the CAD systems to diagnose and classify the breast cancers. neurons are the base of all neural networks including CNN. CNN is designed to learn the latent and intrinsic features from 2D or 3D images in a supervised manner, it is one of the most widely used deep learning techniques, mostly utilized for time series forecasting, natural language processing, and picture classification. State-of-the-art performance in a variety of application fields, including computer vision, image recognition, speech recognition, natural language processing, and speech recognition, has been made possible by its capacity to extract and recognize fine characteristics [15][16][17][18]. Typically, CNN model is composed of multiple layers such as an input layer followed by several convolutional layers, pooling layers and an output layer which includes dense layers. Convolution layer has many neurons that are connected spatially and share the weight and bias. Standard convolution layer converts the input image into a feature map using convolution operation. At convolution layer, the input data is mapped with a group of kernels that produces a new feature map and this process is called convolution. Due to the advent of large-scale training data such as ImageNet [19], CNN displays superior performance in large-scale visual recognition. Researchers are also encouraged to adapt CNNs that have already been trained on ImageNet to other domains and datasets [20], such as breast cancer images datasets, due to the CNN's outstanding performance. Furthermore, CNN typically produces a more discriminative picture representation [21], which is necessary for breast cancer image classification. A pre-trained model is a saved network that was already trained on a large dataset, usually on a large-scale image-classification task. the pretrained model can be used as it is or transfer learning can be used to adapt this model to a specific task.

. The majority of the most advanced CNNs available today can be used for precise image classification. There are multiple CNN architectures such as ImageNet, Inception-v4, ResNet-50, Inception-ResNet, Xception, Inception-v3, Inception-v1, and ResNetXt-50 LeNet-5, AlexNet, GoogleNet, VGGNet [22] such powerful backbones pretrained on such big datasets of extensive categories drawn from a diversity of sources could learn powerful representations with labels. The intuition behind transfer learning for image classification is that if a model is trained on a large and general enough dataset, this model will successfully serve as a generic model of the visual world. You can then make use of these learned feature maps without having to start from scratch by training a large model on a large dataset.

Ensemble learning methods in the decision-making stage is a strong approach in which confidence scores of multiple base learners are grouped together to obtain the final prediction about an input sample. It improves the prediction capability and accuracy of the overall model which its individual base models cannot achieve. This strategy also increases the robustness of the model. Ensemble learning helps to correct the false positive or false negatives when a model makes a biased decision for a specific test sample. It decreases the variance of the prediction errors by adding some bias to the competitive base learners. The most famous ensemble approaches in literature that used in different applications are average probability, majority voting and weighted average probability. In the past, several ensemble-based studies have been made for breast cancer histology image analysis [23]. In our study case, we use totally different approach, in such a way every model is trained fully independently on the whole dataset, it is possible to train any numbers of models such as three or five, these models then will have output, the algorithm will make final prediction based on the average of all the outputs of models it makes on fully independent trained models.

## 2. Literature review

In literature many researches have been made using Artificial intelligence (AI) which is a term used to refer to the creation of models that have intelligent behaviors related to intelligent beings without the human interventions. AI utilizes a wide range of tools and principles such as math, logic, and biology. Modern AI technologies are becoming more and more capable of handling diverse and unstructured data, including images and natural language text, which is an important feature. A topic of which has had great interest is the use of artificial intelligence in medicine. many researchers use machine learning in the cad system, which is considered a branch of AI. Machine learning enables systems to find features and derive their own rules to make automatic decisions and predicts values [24]. ON another hand, deep learning is also used in CAD breast cancer detection systems and in many applications where deep learning can be a three-layer or more neural network. These neural networks make an effort to mimic how the human brain functions, enabling it to "learn" from vast amounts of data. Additional hidden layers can be added to enhance and refine the performance of the neural network model. In this literature review will cover both machine learning and deep learning networks. In [25] article, they used Machine learning algorithms (logistic regression, random forests, support vector machines) where they used number of features such as BMI, Glucose, Insulin, HOMA, Leptin, Adiponectin, Resistin and MCP-1. Using support vector machine (SVM) models has achieved sensitivity (82 % - 88%) and specificity ranging between 85 and 90%. The 95% confidence interval for the AUC was [0.87, 0.91]. In [26] Rahul Karmakar has proposed five classifiers (K-Nearest Neighbors (KNN), Random Forest (RF), Decision Trees (DT), Logistic Regression (LR), and Support Vector Machines (SVM) to measure how they perform on WISCONSIN datasets predicting breast cancer. After using the k-fold cross-validation technique, Random Forest produced the best results. The dataset was divided into 90% training and 10% testing. The score of Random Forest, was nearly 0.96488 using cross-validation, which was the highest among them. It was concluded that compared to other classifiers used in their study, Random Forest is significantly more accurate. However, Logistic Regression performed admirably and provided accuracy that was comparable to Random Forest. The best classifier for predicting breast cancer, according to their analysis of the application of various classifiers Random Forest with different divisions of training data and testing data.

In [27] the work has developed a CAD system using convolutional neural network where they built the model with four inputs to accommodate the four images with different magnification levels in parallel. The model depended on EfficientNet-B0 as the core of the CAD system which it used histopathological images to classify using the neural network and it performed well surpassing machine learning algorithm and some other neural networks.

In [28] VGG16, VGG19, and ResNet50, three popular pre-trained deep CNN models, were used for both full training and transfer learning. Because it is so difficult to categorize breast cancer histology images, a very sophisticated architecture is needed to solve this issue. Due to their more intricate architecture, the VGG16, VGG19, and ResNet50 pre-trained CNN models are extremely preferable. Additionally, these models have demonstrated comparatively high performance for difficult computer vision problems across a variety of domains and it was concluded that pre-trained VGG16 with logistic regression classifier produced the best performance with 92.60% accuracy, 95.65% area under ROC curve (AUC), and 95.95% accuracy precision.

In [29] the authors proposed network with three stages in which the first stage has three parallel CNN branches with deep residual blocks. The next the three parallel CNN branches were merged to build a feature fusion. Finally, the features are classified at the last stage. The Breakhis dataset was used to evaluate this approach using the four magnification factors where it achieved 97.14% accuracy.

Another fusion technique which uses the layers of the pretrained VGG19 which can provide a better initial wight optimization. those fused layers which can approximately cover the nuclei-scale, nuclei organization, and structure-scale features, the robustness of this so-called FCNN is embodied in its ability to cover multi-scale information, while other comparative ones can only focus on a certain level of information, this was proposed by [30].On other hand, mammograms are also used in ensemble approaches, in a way the ensemble classifier and feature weighting algorithm were built where an ensemble classifier model is designed using k-nearest neighbor (KNN), bagging, and eigenvalue classification (EigenClass) to determine whether a mammogram contains normal, benign, or malignant tumors based on a majority voting rules as in [31].

## 3. Proposed Methodology

We looked to the literature and found it was not mentioned about drawing on different well-recognized CNN models ensemble that are previously trained fully and then averaged ensembled. The decisions which are based on the common behaviors of the classifiers, making them more reliable and avoiding overfitting [32]. In our approach we took different popular convolutional neural models, next trains every model independently then combines them in parallel. The image to be classified will be entered to the different models simultaneously to achieve more accuracy and better performance. Every model will predict its classification independently and its influence on the model will be proportional to its accuracy. After the classification of every model, each model will have a probability output between 0 and 1. if the output is less than .5 that means it belongs to the first class and if it is bigger than .5, that means it belongs to the second class. We sum the last outputs of models and multiply every model by its accuracy, then we divide by sum of the models accuracies as shown in the next function. In our approach we used state of the art models such as Inception-V3, ResNet50 and DenseNet-201 after finetuning them to make them suitable for recognition of breast cancer type. The approach can be illustrated in fig. 1.
which shows independently trained models connected in parallel. Each model will have different effect on the model depending on its accuracy according to the following formula.

$$f = \frac{\text{first model output} \times w1 + \text{second model output} \times w2 + \text{third model output} \times w3}{w1+w2+w3}$$

This design will benefit from the strength of every model in way that increases the accuracy of the whole CAD system (automatic breast cancer detection). w represents the accuracies of the models. Gathering those models altogether to use advantages of every model, surely every model will be better at detecting specific lines or edge, and may this model will have defects in detecting and classifying a kind of other edges.

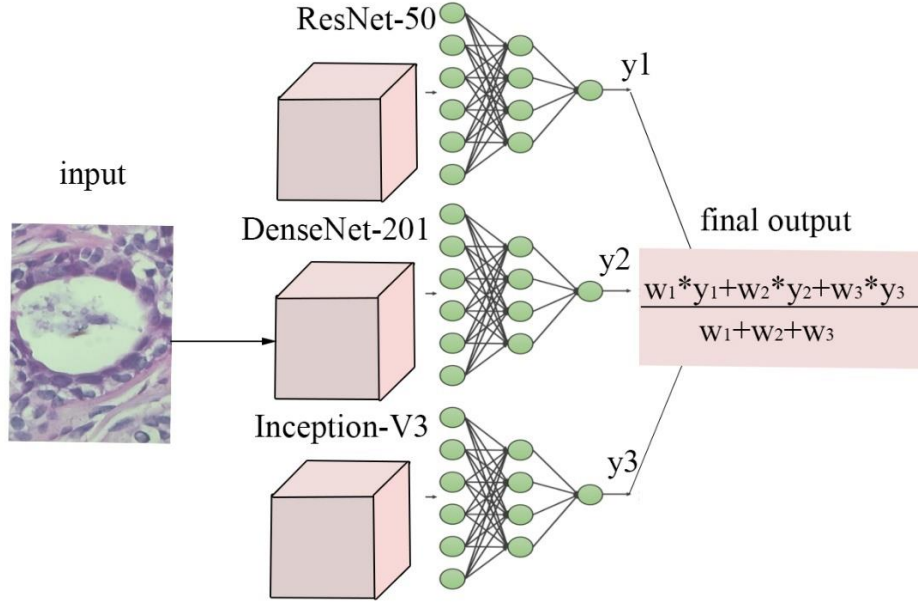

**Fig. 1** weighted ensembled fully trained models

### 3.1. Dataset

actually, there are many datasets which include labeled data that we can use to train and test our models, some of the datasets can be used in convolutional neural networks and other can be used in machine learning. There are leading types of images such histopathological images and mammogram images and many other types, we used histopathological images which are offered by Breakhis dataset. The Breast Cancer Histopathological Image Classification (Breakhis) is composed of 9,109 microscopic images of breast tumor tissue collected from 82 patients using different magnifying factors (40X, 100X, 200X, and 400X). To date, it contains 2,480 benign and 5,429 malignant samples (700X460 pixels, 3-channel RGB, 8-bit depth in each channel, PNG format). This database has been built in collaboration with the P&D Laboratory – Pathological Anatomy and Cytopathology, Parana, Brazil (http://www.prevencaoediagnose.com.br). researchers find this database a useful tool since it makes future benchmarking and evaluation possible. Fig. 2 below shows the histopathological images for both malignant and benign cancer.

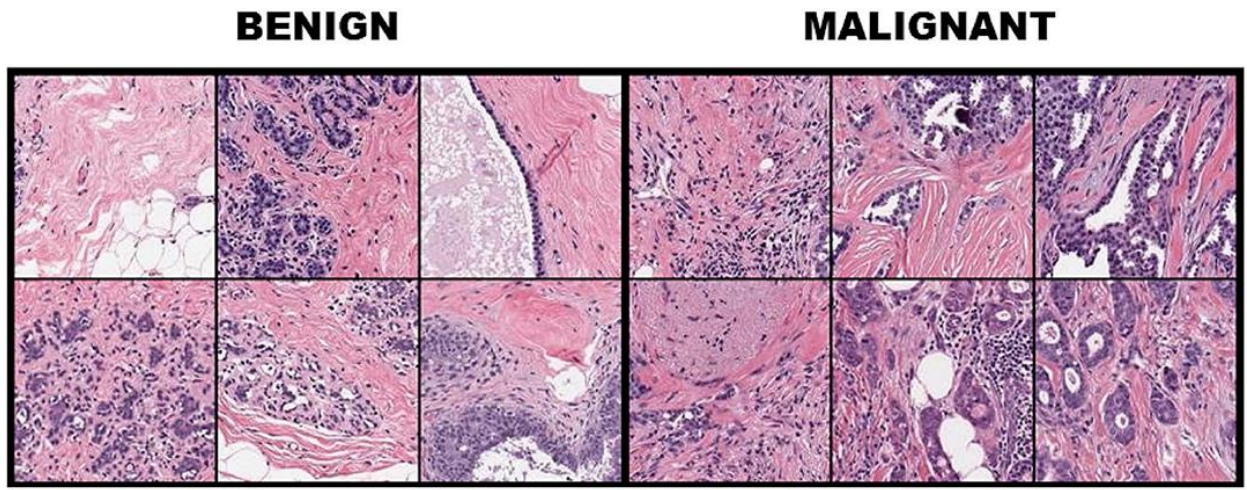

**Fig.** 2 samples of datasets for breast cancer

### 3.2. Convolutional models

*3.2.1. Resnet-50*

ResNet stands for residual network, which points to the residual blocks that make up the architecture of the network. Residual learning framework was presented to ease the training of networks that are considerably deeper than those used in another architectures. the layers are reformulated as learning residual functions with reference to the layer inputs, as alternative to learning unreferenced functions [33]. The ResNet architecture was built in response to a surprising observation in deep learning research which is making neural network deeper was not always improving the results. The ResNet model follows two basic design rules. Firstly, the number of filters in every layer is the same based on the size of the output feature map. Secondly, if the feature map's size is divided by 2, it has double the number of filters to keep the time complexity of every layer. ResNet-50 is made up of 50 layers that are arranged into 5 blocks, each consisting of a set of residual blocks. The residual blocks ease the preservation of information from previous layers, which helps the network to learn better representations of the input data. Bottleneck design was used for the building block in the ResNet-50, A bottleneck residual block uses 1×1 convolutions, known as a "bottleneck", which decreases the number of parameters and matrix multiplications. This allows much faster training of every layer. Also, It uses a stack of three layers instead of two layers in each residual function .

*3.2.2. DenseNet-201*

ResNet has shown that convolutional networks can be considerably deeper, efficient and more accurate to train if they consist of shorter connections between layers near to the input and those near to the output. the Dense Convolutional Network (DenseNet) connects each layer to every other layer in a feed-forward fashion. These kinds of networks are structurally different; however, their basic idea is to use shortcut connections from shallow layers to deep layers. This connection method can circumvent gradient vanishing problem in the networks with deep layers. For each layer, the feature-maps of all earlier layers are used as inputs, and its own feature-maps are used as inputs into all next layers. DenseNets have several appealing advantages: they decrease the vanishing-gradient problem, boosting feature propagation, help feature reuse, and considerably decrease the number of parameters. DenseNets achieved significant improvements over the state-of-the-art on many tasks whilst requiring less computation power to achieve high performance [34]. The architecture is divided into dense blocks with all the successive layers in each block where it uses one-by-one convolution to maintain the spatial resolution, but it reduces the depth of the feature map, followed by max pooling to decrease the feature map size. There are different DenseNets types, such as DenseNet-121, DenseNet-169, DenseNet-201, DenseNet-264, etc., our study used DenseNet-201 which consists of 201 layers with more than 20 M parameters. Fewer parameters are used in DenseNets compared to traditional CNNs because there are no unnecessary feature maps. the structure of DenseNets is divided into dense blocks where the feature map dimensions stay constant inside a block having different filters.

*3.2.3. Inception-v3*

Inception v3 substantially focuses on utilizing less computational power by modifying the previous Inception architectures. The paper "Rethinking the Inception Architecture for Computer Vision "proposed this idea which was published in 2015. It was co-written by Christian Szegedy, Vincent Vanhoucke, Sergey Ioffe, and Jonathon Shlens. In comparison to VGGNet, Inception Networks (GoogLeNet/Inception v1) have demonstrated to be more efficient in computation, both in terms of the number of parameters generated by the network and in terms of memory and other resources which made it more cost-efficient. If any changes in an Inception Network are to be made, care has to be taken to ensure that the computational advantages are achieved. Therefore, the adaptation of an Inception network for different task cases turns out to be a problem because of the uncertainty of the efficiency of the new network. In an Inception v3 model, several techniques for optimizing the network have been suggested to ease the constraints for easier model adaptation. The techniques involve dimension reduction, regularization, parallelized computations and factorized convolutions.

### 3.3. Preprocessing

Preprocessing techniques are an important part of the deep learning process. By carefully preprocessing data, we can improve the performance of our trained model and make it more robust to variations in the input data. Also, it can have a significant impact on the performance of the trained model. There are a variety of preprocessing techniques that can be used in deep learning, depending on the specific dataset and task at hand. Some common preprocessing techniques that we used are as the followings:

*3.3.1. Normalization:*
If the input images are normalized, the model will converge faster and more accurate. When the input images are not normalized, the shared weights of the network have different calibrations for different features, which can force, the time of cost function to converge, taking longer time and in less proficiently way. Normalizing the data makes the cost function much easier to train.

*3.3.2. Horizontal flip*
it is a type of transformation that is used in data augmentation techniques to increase the dataset used in deep learning that causes the images to flip horizontally from left to right. It makes a mirrored image of the original image along the vertical axis.

*3.3.3. Vertical flip*
A vertical flip is the transformation of a geometric figure or image in which every point is reflected across a horizontal axis. This means that the top of the figure or image becomes the bottom, and the bottom becomes the top. Vertical flips are often used in image processing. They can be used to create mirror images or to simply invert an image. Vertical flips can also be used as a data augmentation technique in machine learning to increase the size and diversity of a training dataset.

*3.3.4. Shear*
Shear in data augmentation is a geometric transformation that skews the image along a particular axis. This can be used to create a more diverse training dataset for machine learning models, and to help them learn to generalize to new data. There are two types of shears: horizontal shear and vertical shear.
Horizontal shear: skews the image to the left or right.
Vertical shear skews the image up or down.
The amount of shear is typically specified by a shear angle, which is measured in degrees. A shear angle of 0 degrees means no shear, while a shear angle of 45 degrees means that the image is sheared at a 45-degree angle.

*3.3.4. Zoom*
Zoom data augmentation is a technique used to increase the size and diversity of a training dataset for machine learning models by zooming in or out on images. This can help models learn to be more robust to changes in scale, and to generalize better to new data. Zoom data augmentation can be implemented using a variety of methods. One common approach is to use a random zoom factor, which can be specified as a range or a single value. For example, a zoom factor of 0.5-1.5 would randomly zoom the image in or out by a factor of between 0.5 and 1.5. Another approach to zoom data augmentation is to use a fixed zoom factor. This is typically used when you want to zoom in on a specific region of the image. For example, you might want to zoom in on the face of a person in an image to help a model learn to recognize faces.

## 4. Training methodology

We used TensorFlow for training the models on the laptop with these specifications: GPU Nvidia Quadro P600 which has 4 giga bit ram, and the CPU is Core I7 with ram 16 Giga bit. The back-forward optimization method was Adam with learning rate 0.0001. Furthermore, we used early stop method that will stop the training if there are not any improvements after five epochs. The global average pooling layer is used which averages every feature map and provides a single value. It is followed by a dense layer which is followed by dropout with a probability of 0.4. the final layer is the classification layer which is consisted of only one neuron with sigmoid activation function.

**Results**

In this study, we are going to show the effectiveness of using weighted ensemble approach in increasing the metrics of evaluation such as increasing accuracies, F1-scores and recall and decreasing the false positives and false negatives predictions. The scikit-learn learn methods have been used to measures the accuracies and other evaluation metrics. We have made various experiments to do the binary classification of breast cancer histopathology images. We have analyzed the performance of state-of-the-art CNN models and have shown the performance metrics of each model alone, and also a comparative analysis with the weighted averaged ensembled one. The weighted averaged ensembled model increased the accuracy by almost one percent, and it also increased all the evaluation metrics. Accuracy, F1 score, recall, and precision are chosen as the evaluation metrics. Firstly, we show the independent models metrics which were the results of using classification report method in scikit learn. Fig. 3 shows the confusion matrix of the ResNet50. table 1 shows the performance metrics of the ResNet-50. DenseNet-201 confusion matrix is show on Fig. 4. Fig. 5 and table 3 shows the evaluation metrics of the Inception-V3 model. Fig. 6 demonstrates the confusion matrix of the weighted averaged ensembled model which shows that the number of false positive and false negative has decreased and it is better than all the models and it also increased the accuracy of the whole model by almost 1 percent. The final accuracy is 98% while the best independently model is 97%.

**Table** 1 for the metrics of ResNet-50

| Type of metric | precision | recall | F1-score | support |
|---|---|---|---|---|
| 0 | .96 | .96 | .96 | 372 |
| 1 | .98 | .98 | .98 | 815 |
| Macro avg | .97 | .97 | .97 | 1187 |
| Weighted avg | .98 | .98 | .98 | 1187 |

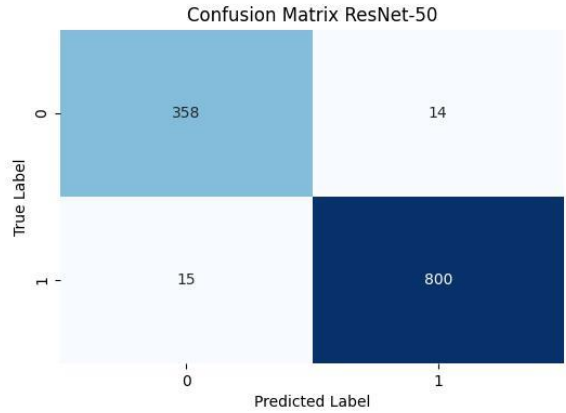

**Fig.** 3 confusion matrix for ResNet-50

**Table** 2 DenseNet-201 metrics results

| Type of metric | precision | recall | F1-score | support |
|---|---|---|---|---|
| 0 | .89 | .94 | .91 | 372 |
| 1 | .97 | .95 | .96 | 815 |
| Macro avg | .93 | .94 | .94 | 1187 |
| Weighted avg | .95 | .95 | .95 | 1187 |

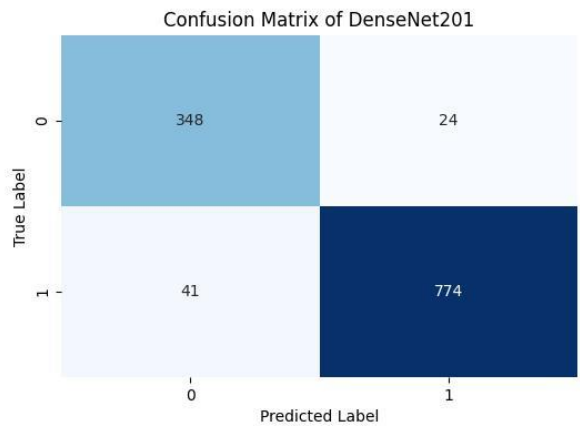

**Fig.** 4 confusion matrix for DenseNet-201

**Table** 3 inception-v3 metrics

| Type of metric | precision | recall | F1-score | support |
|---|---|---|---|---|
| 0 | .94 | .96 | .95 | 372 |
| 1 | .98 | .97 | .98 | 815 |
| Macro avg | .96 | .96 | .96 | 1187 |
| Weighted avg | .97 | .97 | .97 | 1187 |

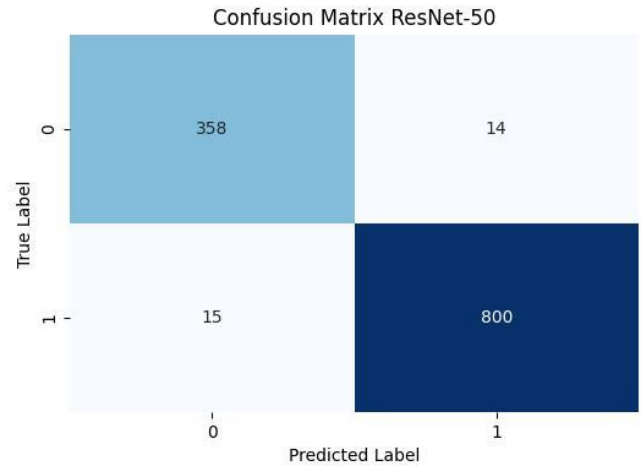

**Fig.** 5 confusion matrix for Inception-V3

The three state-of-the-art models trained on Breakhis are used to perform experiments. accuracy_score method in scikit learn is used to measure the accuracies. One of These models is ResNet-50 which achieved the best accuracy of 97.55% on the whole dataset. While the accuracy of Inception_V3 on the whole dataset is 96.63% on BreakHis dataset. The lowest performance, in this case, is observed by the DenseNet201 model 94.55 Accuracy. precision, recall, and F1-score of all models on the whole dataset are shown in above Tables 1,2 and 3. In Table 6, the results for complete BreakHis dataset are given. Table 6 shows the evaluation metrics of the weighted averaged ensembled.

**Table** 4 weighted averaged ensembled model metrics

| Type of metric | precision | recall | F1-score | support |
|---|---|---|---|---|
| 0 | .96 | .98 | .97 | 372 |
| 1 | .99 | .98 | .99 | 815 |
| Macro avg | .98 | .98 | .98 | 1187 |
| Weighted avg | .98 | .98 | .98 | 1187 |

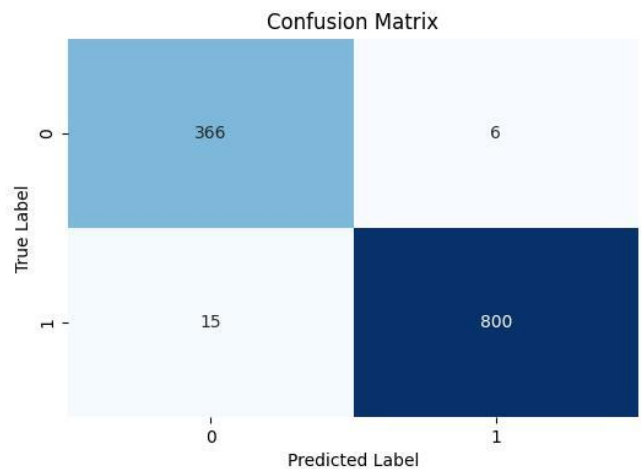

**Fig.** 6 confusion matrix for the weighted averaged ensembled model

## 5. Conclusion

If any outputs of any trained models were weighted summed, then averaged by the sum of the accuracies, their final output accuracy and other evaluation metrics will be better than the best of the joining model. We tried the ensemble techniques on many models and dataset and the final output was always better. This result is because every model will share its experience and its expressiveness and every model will have different defects than others, in this way every model will correct the mistake of other models using the weighted average ensemble. Also, we tried to average them in a way that there is not any weight influence on every model, however this approach did not have big impact on the final output, the F1-score has decreased by one percent from the weighted average method.

**Conflict of Interest**

The authors declare that there is no conflict of interests regarding the publication of this article

**Acknowledgment**

This work was supported in part by National Natural Science Foundation of China (No. 62073120), the Natural Science Foundation of Jiangsu Province (No. BK20201311).

Type equation here.